\UseRawInputEncoding
\pdfoutput=1
\RequirePackage{fix-cm}
\documentclass{svjour3}         %onecolumn(standard format)
\smartqed  % flush right qed marks, e.g. at end of proof
\usepackage{graphicx}
\usepackage{epstopdf}
\usepackage{geometry}
\usepackage[numbers,sort&compress]{natbib}
\usepackage{bm}
\usepackage{amsmath,amssymb,amsfonts}
\usepackage{algorithm}
\usepackage[noend]{algpseudocode}
\usepackage{graphicx}
\usepackage{multirow}
\usepackage{etoolbox}
\usepackage{verbatim}
\usepackage{lineno}
\usepackage{setspace}
\usepackage{textcomp}
\usepackage[tight]{subfigure}
\usepackage{caption}
\usepackage{subfig}
\usepackage{epstopdf}
\usepackage[misc]{ifsym} % tongxunzuozhetubiao
\usepackage{booktabs}

\begin{document}
\title{Multi-stage Progressive Reasoning for Dunhuang Murals Inpainting %\thanks{Grants or other notes
}


\author{Wenjie Liu\textsuperscript{1}         \and
        Baokai Liu\textsuperscript{1}        \and
        Shiqiang Du\textsuperscript{1,2}      \and
        Yuqing Shi\textsuperscript{3}         \and
        Jiacheng Li\textsuperscript{1}        \and
        Jianhua Wang\textsuperscript{1}
             %etc.,~\Letter
}

\institute{%
	\begin{itemize}
		\item[\textsuperscript{\Letter}] {Shiqiang~Du } \\
		%		Tel.: +123-45-678910 \\
		%		Fax: +123-45-678910 \\
		\email{shiqiangdu@hotmail.com}
		\at
		\item[\textsuperscript{1}] Key Laboratory of China's Ethnic Languages and Information Technology of Ministry of Education, Chinese National Information Technology Research Institute, Northwest Minzu University, Lanzhou, Gansu, 730030 China
		\at
		\item[\textsuperscript{2}] College of Mathematics and Computer Science, Northwest Minzu University, Lanzhou, Gansu, 730030 China
      	\at
		\item[\textsuperscript{3}] College of Electronic Engineering, Northwest Minzu University, Lanzhou, Gansu, 730030 China
	\end{itemize}
}

%\institute{F. Author \at
 %             first address \\
  %            Tel.: +123-45-678910\\
   %           Fax: +123-45-678910\\
    %          \email{fauthor@example.com}           %  \\
     %      \and
      %     S. Author \at
       %       second address
%}

\date{Received: date / Accepted: date}

%\twocolumn[
%\begin{@twocolumnfalse}
\maketitle
\begin{abstract}
Dunhuang murals suffer from fading, breakage, surface brittleness and extensive peeling affected by prolonged environmental erosion. Image inpainting techniques are widely used in the field of digital mural inpainting. Generally speaking, for mural inpainting tasks with large area damage, it is challenging for any image inpainting method. In this paper, we design a multi-stage progressive reasoning network~(MPR-Net)~containing global to local receptive fields for murals inpainting. This network is capable of recursively inferring the damage boundary and progressively tightening the regional texture constraints. Moreover, to adaptively fuse plentiful information at various scales of murals, a multi-scale feature aggregation module~(MFA) is designed to empower the capability to select the significant features. The execution of the model is similar to the process of a mural restorer~(i.e., inpainting the structure of the damaged mural globally first and then adding the local texture details further). Our method has been evaluated through both qualitative and quantitative experiments, and the results demonstrate that it outperforms state-of-the-art image inpainting methods.

\end{abstract}
\section{Introduction}
\label{sec:Introduction}
\begin{figure*}[t]% h ?????t?? ;b???p ???; ???[tbp]		
	\centering
	\subfigcapskip=4pt % ?????????????
	\subfigure[Discoloration]{
		\begin{minipage}{0.22\linewidth}	
			\centering
			\includegraphics[width=1.05\linewidth]{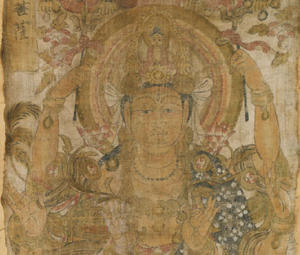}
		\end{minipage}
	}
	\subfigure[Erosion]{
		\begin{minipage}{0.22\linewidth}	
			\centering
			\includegraphics[width=1.05\linewidth]{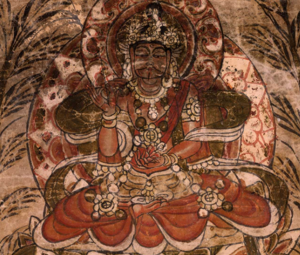}
		\end{minipage}
	}
	\subfigure[Breakage]{
		\begin{minipage}{0.22\linewidth}	
			\centering
			\includegraphics[width=1.05\linewidth]{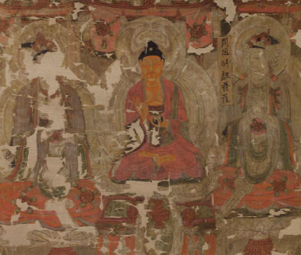}
		\end{minipage}
	}
	\subfigure[Abscission]{
		\begin{minipage}{0.22\linewidth}	
			\centering
			\includegraphics[width=1.05\linewidth]{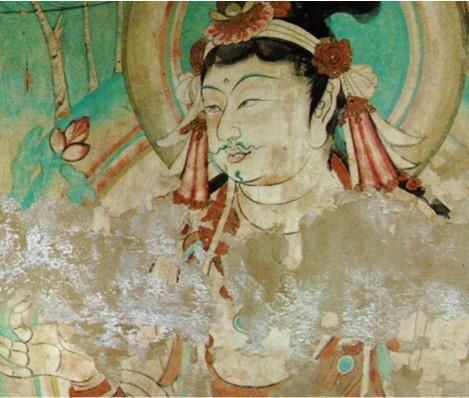}
		\end{minipage}
	}
	\hspace{0.1 cm}
	\caption{Murals of different damage types.}
	\label{sunhuai}
\end{figure*}

Dunhuang murals refer to the paintings on the interior walls of the Dunhuang Caves in China, renowned for their superb technique and magnificent scope, which are valuable material cultural heritage. Dunhuang murals have profound historical origins, spanning multiple dynasties, including the Western Wei, Northern Zhou, Tang, and Song dynasties, and experienced more than a thousand years of historical vicissitudes. However, the degradation of many murals can be attributed to natural elements such as radiation, sand, dust, humidity, and harsh temperatures, which have caused them to become damaged and lose their original quality, and the different classifications of deteriorated murals are presented in Fig.~\ref{sunhuai}. As cultural relics are non-renewable resources that will eventually die out with time, This is a severe loss of human civilization. Therefore, it is significant to explore effective inpainting techniques for murals.

\begin{figure}[htbp]%
	\centering
	\includegraphics[width=0.95\textwidth]{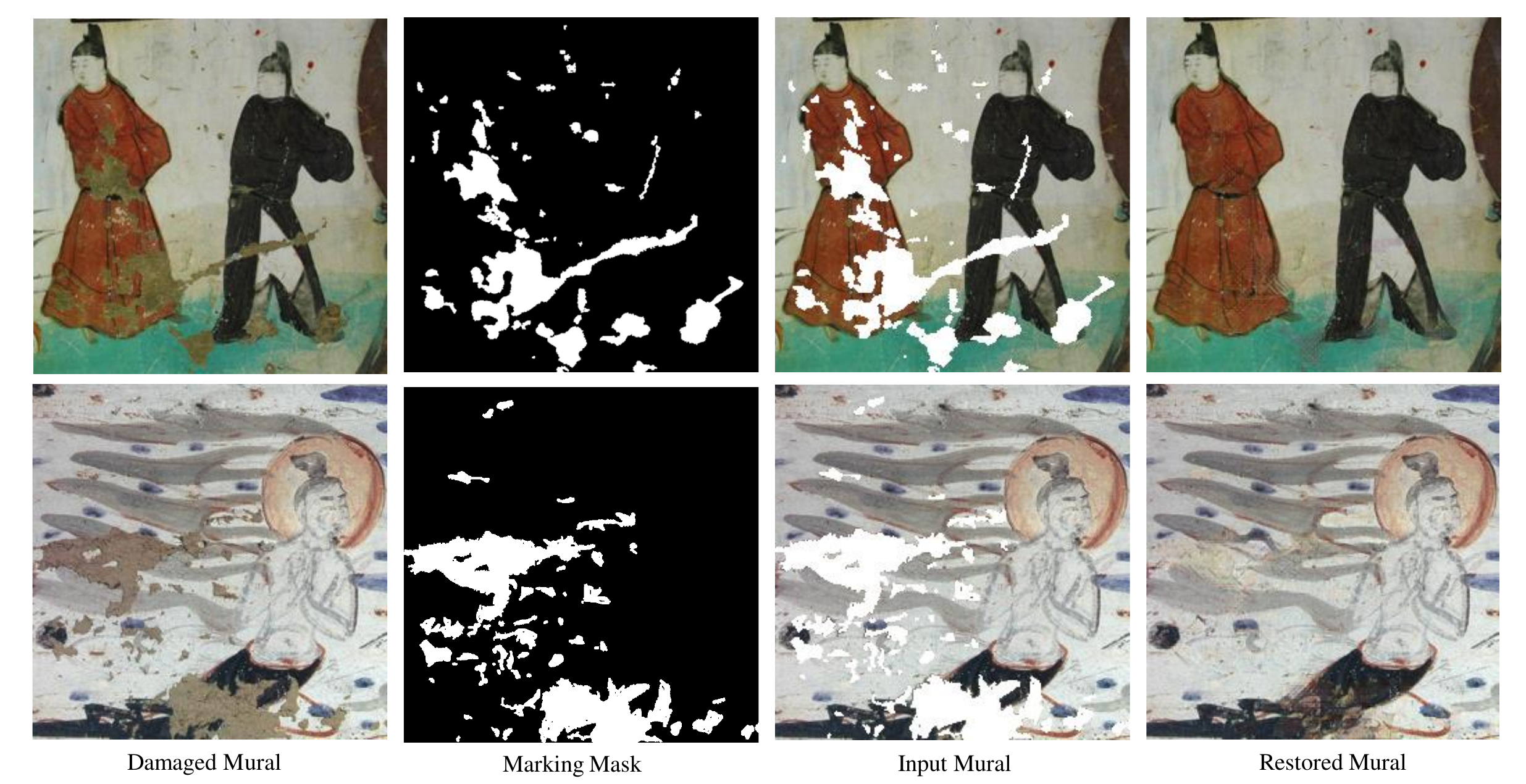}   %./figure/framework.eps
	\caption{The application of inpainting damaged murals.}
    \label{real}
\end{figure}

In the past, people used manual inpainting to fill in missing areas of murals, but it has several drawbacks. Firstly, it's risky because it can damage the mural surface. Secondly, it's time-consuming and requires a high level of skill. Thirdly, it's costly and irreversible. As a result, the cultural heritage community is now more careful about using manual inpainting and is exploring other options like digital inpainting, which uses computer software to recreate missing areas. Digital inpainting is less risky, more efficient, and can be corrected easily, so it's gaining popularity in the restoration of cultural heritage sites. In this field, Pan et al. \cite{Pan2003Digital}~first proposed a set of technical solutions for digital mural inpainting from a macro perspective and described its system architecture and operation mechanism. Moreover, Flusser et al.~\cite{B2003Image} combined chemical analysis in the mural inpainting process. Wang et al. \cite{2011Virtual} designed the semantic learning inpainting framework for broken human faces in ancient murals. Criminis et al. \cite{2003Object} introduced the texture inpainting algorithm based on sample block matching. Yang et al. \cite{Tianshui2011Dunhuang} improved the priority function of the Criminisi inpainting algorithm using D-S theory with a data fusion approach and researched the inpainting of color changes and human-contaminated damages in Dunhuang murals. Li et al. \cite{2018A} defined structure-first and texture-later inpainting strategies to restore broken murals through human-computer interaction digitally. Liu et al. \cite{liu2022dunhuang} designed a novel edge detector based on self-attention combined with convolution to generate line drawings of Dunhuang murals. Deep learning-based image inpainting methods have made significant progress in recent years, but they may face difficulties when it comes to restoring murals. This is because the process of painting murals involves creating smooth lines and evocative colors that are different from those found in natural photographs. Therefore, image inpainting methods that are designed for photographs may not be well-suited for restoring the unique edge structure and texture details of a mural. The network splits the inpainting of the mural into two articulated subtasks, which can infer the damaged structure and texture from the perspective of global and local receptive fields. In addition, we construct the Dunhuang murals dataset and real masks dataset, where the real masks are extracted from the broken regions of the damaged murals. Therefore, our model can realize practical breakages of the mutilated murals owning a higher application in specific mural inpainting. The inpainting performance of our method on realistic damaged murals is shown in Fig.~\ref{real}.

In summary, the main innovations and contributions of this paper are as follows:
\begin{enumerate}
     \item The multi-stage progressive reasoning network is designed with global to local receptive fields to solve the problem of the single semantic in recurrent networks, which achieves both semantically reasonable structures and detail-rich textures.
     \item To effectively utilize the features inferred from the model, a multi-scale feature aggregation module is designed to empower the capability of dynamic selection from the significant semantics, leading to semantic consistency results with adaptive features.
     \item We analyze the model in terms of both qualitative and quantitative, which demonstrate the superiority compared with several state-of-the-art methods in the Dunhuang murals and benchmark datasets.
\end{enumerate}	
\section{Related work}
Image inpainting aims to fill the defective areas with realistic content and ensure consistency regarding global structures and texture details. With the continuous advancement of research work over the years, significant progress has been made in image inpainting. Current image inpainting methods can be divided into traditional diffusion methods and deep learning-based methods.

$\textbf{Traditional inpainting methods:}$~The diffusion-based approach was pioneered by~Bertalmio~et al. \cite{bertalmio2000image} used smooth local information to constrain the inpainting results. Ballester et al. \cite{ballester2001filling} transformed the image inpainting task into a variable differential problem in terms of the gradient and gray-scale of the image. Tschumperl¨¦~and~Deriche \cite{tschumperle2005vector} presented a general anisotropic diffusion method with better inpainting results. However, these methods cannot handle larger missing regions due to the limitations of boundary expansion prediction. The patch-based image inpainting approaches such as \cite{James2008Scene,sun2005image} typically propagated appearance information from the remained regions or other source images into the missing regions through various manually defined similarity metrics between patches. Chen et al. \cite{chen2023multi} used multi-scale generative adversarial networks with edge detection for image inpainting. Since these methods rely on low-dimensional feature information, they are difficult to achieve satisfactory results in stable semantic tasks.

$\textbf{Deep learning-based inpainting methods:}$~From the network design perspective, there can be roughly classified into three types: one-stage, two-stage, and recurrent networks.

For the one-stage networks, Pathak~et al. \cite{pathak2016context} designed a codec architecture trained by pixel reconstruction loss and adversarial loss. To improve the consistency of image complementation, Iizuka~et al. \cite{iizuka2017globally} introduced global and local discriminators to constrain the inpainting results. Inspired by several existing approaches \cite{song2018contextual,yan2018shift}, which own innovative model architecture. Zeng~et al. \cite{zeng2019learning} presented the pyramid-context encoder network to accomplish image complementation by attention transfer. Huang et al. \cite{huang2022region} proposed a novel region-aware attention module to avoid misleading invalid information in holes. Due to the limitations of one-stage architecture lacking sufficient constraints, they occasionally suffer semantic bias and texture-blurring results.

For the two-stage networks, Yu~et al. \cite{yu2018generative} raised an improved generative inpainting network consisting of a coarse network and a refinement network. A fuzzy inpainting result is obtained from the coarse network, and the long-term relevance dependence is then simulated in the refinement network by a contextual attention mechanism. Liu~et al. \cite{liu2019coherent} used a fined deep generative model-based approach with a novel coherent semantic attention layer, which can preserve contextual structures and make more effective predictions of missing parts. Inspired by \cite{liu2019image}, which presented an adaptive feature update method, Yu~et al. \cite{yu2019free} promoted irregular mask update by introducing gated convolution. Nazeri~et al. \cite{2019EdgeConnect} proffered an edge-guided two-stage image inpainting method by restoring the missing edge map and then combining this edge map with the incomplete image as the second stage of the input. Due to the limited structure guidance of edge images, Ren et al. \cite{ren2019structureflow} used smoothed image edges as the structure representation. Guo et al. \cite{guo2021image} proposed a novel two-stream network for image inpainting, which models the structure-constrained texture synthesis and texture-guided structure reconstruction in a coupled manner. Although the two-stage networks can achieve better inpainting results, it is prone to the problem of large training fluctuations and unstable convergence due to their overly deep network structure.

For recurrent inpainting networks, Xiong et al. \cite{xiong2019foreground} and Nazeri et al. \cite{2019EdgeConnect} filled images with contour edge completion and image completion in a step-wise manner to ensure structural consistency, Zhang et al. \cite{zhang2018semantic} divided the process of image inpainting into four different stages and used~LSTM architectures \cite{hochreiter1997long} to control the information flow of the progressive process. However, those methods cannot handle irregular holes common in real-world applications. To address this limitation, Guo~et al. \cite{guo2019progressive} projected a full-resolution residual network with multiple expansion modules. Li et al. \cite{li2019progressive} progressively reconstructed the visual structure, entangling visual feature reconstruction in image inpainting. Li et al. \cite{li2020recurrent} followed a recursive framework in the feature space and designed a recurrent feature inference network with consistent attention. Zeng~et al. \cite{zeng2020high} designed an iterative confidence-based inpainting method. Guo et al. \cite{2019Progressive} offered a full-resolution residual network to fill irregular holes in the original size images. Oh et al. \cite{2019Onion} proffered the onion-peel networks for video completion progressively, enabling it to exploit richer contextual information for the missing regions at every step. However, the reasoning process of previous recurrent methods generally occurs in inflexible network architecture lacking awareness of semantics in different fields. In this paper, a multi-stage progressive reasoning network is designed to satisfy the requirements from the perspective of global structure and local texture, which effectively solves this problem.
\section{Proposed method}
\begin{figure}[htbp]%
	\centering
	\includegraphics[width=\textwidth]{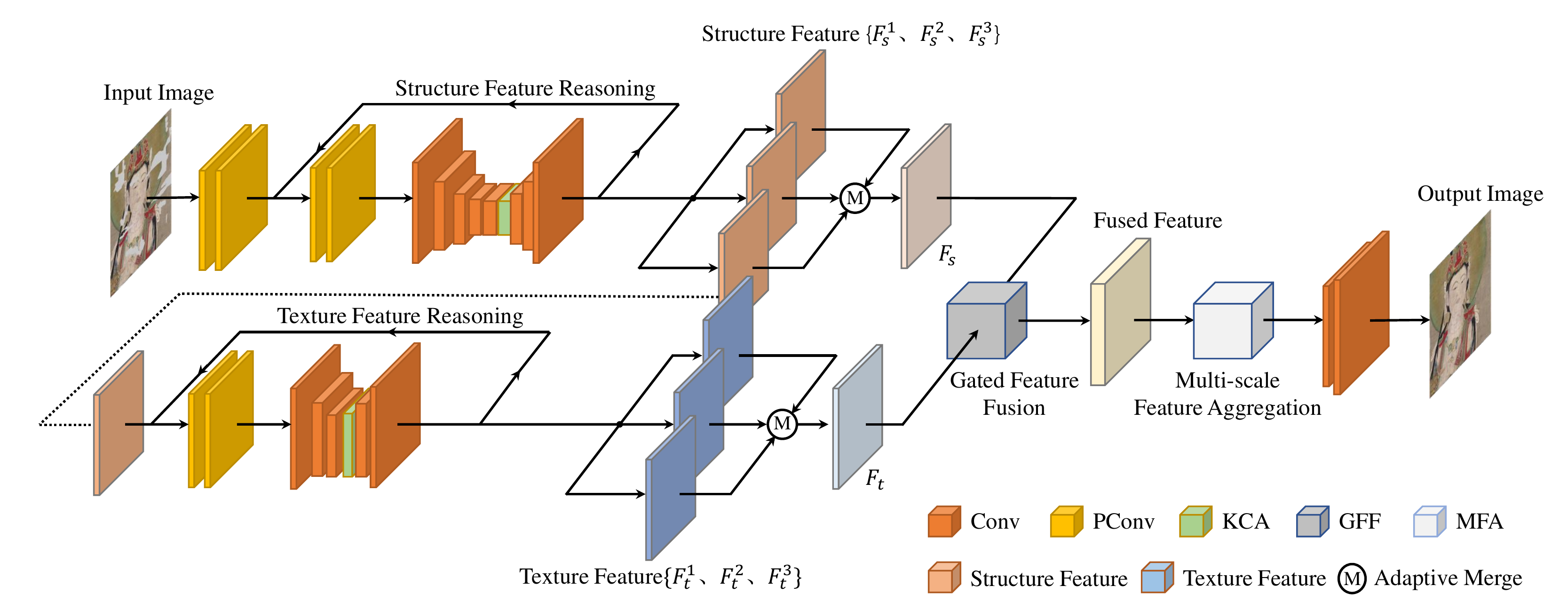}   %./figure/framework.eps
	\caption{The illustration of the proposed MPR-Net. Structure Feature Reasoning and Texture Feature Reasoning are designed to infer missing content progressively. After several times of reasoning, structure and texture feature maps are merged adaptively and then sent to the GFF module for feature interaction. The fused feature is aggregated in multi-scale and then deconvoluted to output.}
    \label{model}
\end{figure}

The design of MPR-Net is inspired by the inference of LG-Net \cite{quan2022image} that the global receptive field is more effective in restoring the structure of an image. In contrast, the local receptive field focuses on texture detail. According to this conclusion, the model is designed with global to local receptive fields based on recurrent architecture, as shown in Fig.~\ref{model}. In this section, we first introduce the multi-stage progressive reasoning~(MPR)~module, then describe the gated feature fusion~(GFF)~and multi-scale feature aggregation~(MFA)~methods, and finally present the objective function of the network.

\textbf{\subsection{Multi-stage progressive reasoning module}}
Compared to natural photographs, murals belong to human-drawn images with unique distinct edges and rich details~(i.e., structure and texture features). The MPR module is designed to model the distinctive semantics of the murals. Specifically, our MPR module can be divided into three parts:
\vspace{0.1cm}
\begin{enumerate}
 \item[a.] A multi-stage progressive network containing global to local receptive fields for dedicated inpainting of unique attributes of the murals.

 \item[b.] An attention mechanism of knowledge consistency establishes the connection for attention scores during recurrent inference.

 \item[c.] A feature merging operator for adaptively weighting structure and texture features.
 \end{enumerate}

Inside the module, The attention mechanism acts with every recursive inference. After filling the damage of the murals, all features generated during inference are merged with adaptive weighting. In the following, we elaborate on the execution flow of the~MPR~module.
\textbf{\subsubsection{Multi-stage progressive network}}
The network consists of two progressive subtasks, where the Structure Feature Reasoning (SFR) owns a global receptive field that comprehensively considers the structure features; the Texture Feature Reasoning (TFR) further restores the missing texture details under the local receptive field. The partial convolution \cite{liu2018image} is introduced to determine the region which needs to be updated in the recurrence:
\begin{equation}
f_{x, y, z}^{*}=\left\{\begin{array}{ll}
W_{z}^{\mathsf{T}}\left(f_{x, y} \odot m_{x, y} \frac{\operatorname{sum}(1)}{\operatorname{sum}\left(m_{x, y}\right)}\right)+b, & \text { if } \operatorname{sum}\left(m_{x, y}\right) !=0 \\
0, & \text { else }
\end{array}\right.
\end{equation}
where $f^*$~denotes the feature map generated by the partial convolution layer. $f^*_{x,y,z}$~denotes the feature values at~$x,y$~locations in the~$z^{th}$~channel. $W_z$~is the~$z^{th}$~convolution kernel of this layer. $f_{x,y}$~and~$m_{x,y}$~are the feature and mask centered at~$x,y$ respectively, and $\frac{\operatorname{sum}(1)}{\operatorname{sum}\left(m_{x, y}\right)}$ is the scale factor. The output result is adjusted when the number of convolution effective input pixels changes. Similarly, the updated mask value at locations~$i, j$~generated by this layer can be formulated as:
\begin{equation}
m_{x, y}^{*}=\left\{\begin{array}{ll}
1, &  \text{if}~\operatorname{sum}\left(m_{x, y}\right) !=0 \\
0, &  \text{else}
\end{array}\right.
\end{equation}

According to the above formula, a new mask with smaller holes can be received after partial convolution of the layers. The difference between the updated and input masks is defined as the area to be repaired in this recurrence. The updated mask remains original until the subsequent iterations.

\textbf{\subsubsection{Knowledge consistent attention}}
In the image inpainting task, the attention mechanism \cite{li2020recurrent} can search for features highly relevant to the match from known regions to fill the missing regions to ensure the semantic consistency of the inpainting results. In particular, we calculate the similarity between known and missing region blocks by cosine distance.
\begin{equation}
{{sim}}_{x, y, x^{\prime}, y^{\prime}}^{i}=\left\langle\frac{f_{x, y}}{\left\|f_{x, y}\right\|}, \frac{f_{x^{\prime}, y^{\prime}}}{\left\|f_{x^{\prime}, y^{\prime}}\right\|}\right\rangle
\end{equation}
where~${{sim}}_{x, y, x^{\prime}, y^{\prime}}^{i} $~denotes the correlation between the masked region pixel at position~$(x,y)$~and the available pixel at position~$(x^{\prime}, y^{\prime})$~in the~$i^{th}$~recurrent. The relevance score of the pixel at position~$(x,y)$ is normalized using the~softmax~function, and the attention score map obtained for that pixel is denoted as:
\begin{equation}
{score}^{\prime}{ }_{x, y, x^{\prime}, y^{\prime}}^{i}=\frac{\exp\left({sim}^{}{ }_{x, y, x^{\prime}, y^{\prime}}^{i}\right)}{\sum_{j=1}^{N} \exp\left({sim}^{}{ }_{x, y, x^{\prime}, y^{\prime}}^{i}\right)}
\end{equation}

To better fit the designed recursive architecture and weaken the discreteness caused by the independent attention computation in each recurrent, adaptive weight values $ \lambda$ are used to connect two adjacent recurrences. Specifically, suppose the pixel at position~$(x,y)$~is valid in the last recurrent. In that case, we adaptively combine the final score of the pixel in the previous recursion with the score computed in this recursion, as follows:
\begin{equation}
{score}_{x, y, x^{\prime}, y^{\prime}}^{i}=\lambda {score}^{\prime}{ }_{x, y, x^{\prime}, y^{\prime}}^{i}+(1-\lambda) {score}_{x, y, x^{\prime}, y^{\prime}}^{i-1}
\end{equation}
if the pixel value at~$(x, y)$~in the previous recursion is invalid, the attention score obtained in this recursion is the final score:
\begin{equation}
{score}_{x, y, x^{\prime}, y^{\prime}}^{i}={score}^{\prime}{ }_{x, y, x^{\prime}, y^{\prime}}^{i}
\end{equation}

Finally, the feature map is reconstructed using the attention scores, and the new feature map of $(x, y)$~points is calculated as follows:
\begin{equation}
\widehat{f}_{x, y}^{i}=\sum_{x^{\prime} \in 1, \ldots W, y^{\prime} \in 1, \ldots H} {score}_{x, y, x^{\prime}, y^{\prime}}^{i} f_{x^{\prime}, y^{\prime}}^{i}
\end{equation}
after the feature map reconstruction is completed, the input features~$F$~are concatenated with the reconstructed feature map~$\hat{F} $~and sent into the pixel convolution layer to obtain the output of this module:
\begin{equation}
{F^{\prime}}^{i}=\phi(|\widehat{F}, F|).
\end{equation}

\textbf{\subsubsection{Adaptive merging operator}}
Since the structure features in early reasoning fully utilize information from the damaged region boundary, which should be more deterministic, in contrast, the inference of texture features lacks sufficient deterministic constraints since nearing the damaged center, which should own more degrees of freedom. Therefore, we design an adaptive merging operator to control the weight, and the~$\beta$~is a learnable parameter. The merged structure features $\boldsymbol{F}_s $ and texture feature $\boldsymbol{F}_t$ are represented as:
\begin{equation}
\boldsymbol{F}_s =\beta \cdot \frac{\sum_{i=1}^{N} f_{x, y, z}^{i}}{\sum_{i=1}^{N} m_{x, y, z}^{i}}
\end{equation}

\begin{equation}
\boldsymbol{F}_t =(1-\beta) \cdot \frac{\sum_{i=1}^{M} f_{x, y, z}^{i}}{\sum_{i=1}^{M} m_{x, y, z}^{i}}
\end{equation}
where~$f^i$~as the~$i^{th}$~feature map generated by the~MPR~module, and $f_{x,y,z}$~as the value of the~$x,y,z$~position in the feature map~$f$. $m^i$~is the binary mask of the feature map~$f^i$, $N$ and $M$ are the numbers of recurrent for SFR and TFR, respectively.

\begin{figure}[htbp]%
	\centering
	\includegraphics[width=0.95\textwidth]{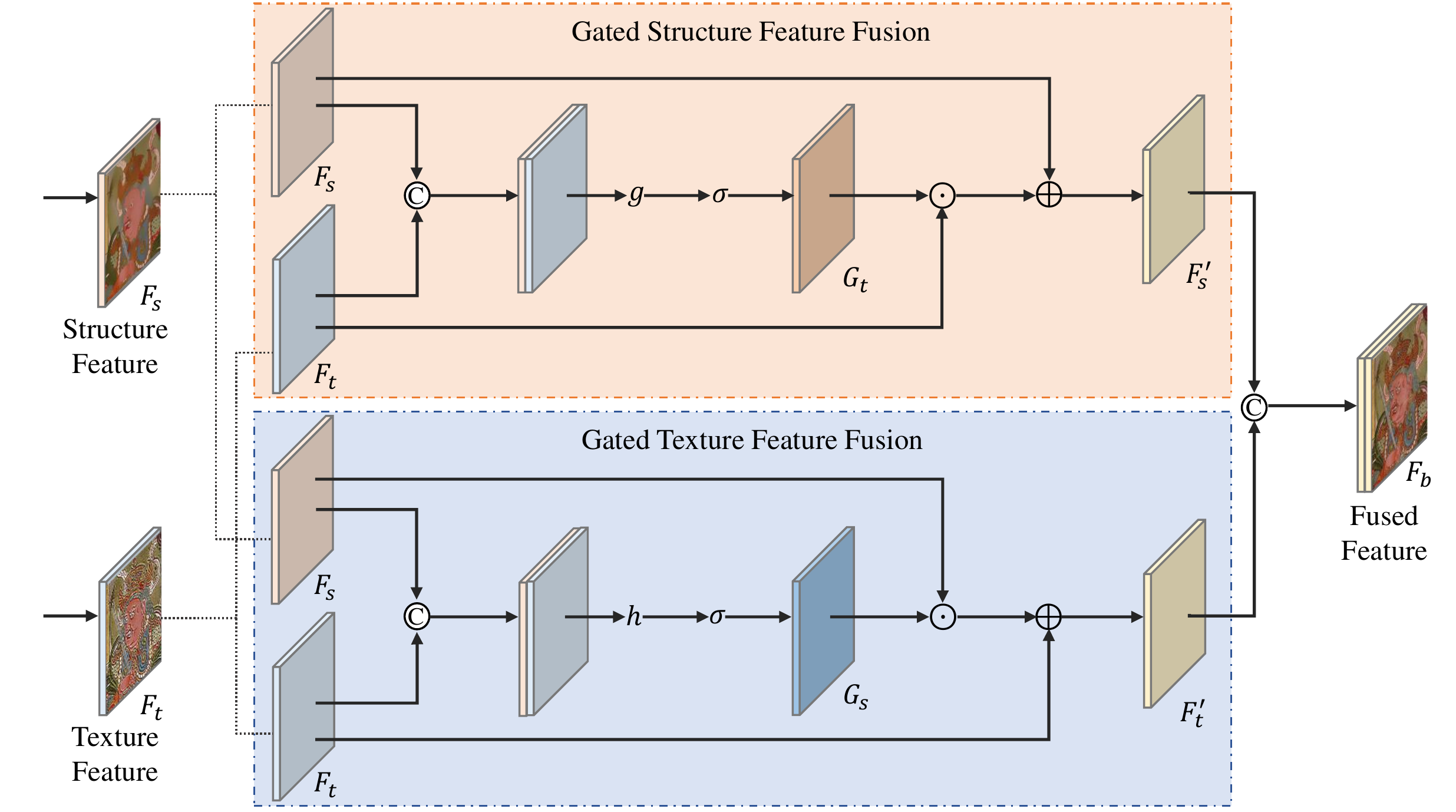}   %./figure/framework.eps
	\caption{Illustration of the Gated Feature
Fusion (GFF) module, which implements interaction between structure and texture features to refine the results.}
    \label{bigff}
\end{figure}

\textbf{\subsection{Gated feature fusion}}
Gated feature fusion module \cite{guo2021image} is introduced to further fuse the inferred structure and texture features, where the soft gating is used to learn weight rationing from a fully exchanged of the two feature information. The redundant information brought by multiple iterations can be discarded based on the fusion operation under feature awareness. The execution flow of the GFF~method is shown in Fig.~\ref{bigff}.

Especially, to construct the texture-aware structure features, we denote the soft gating~$\boldsymbol{G}_t$~for texture information as:
\begin{equation}
\boldsymbol{G}_{t}=\sigma\left(g\left( { Concat }\left(\boldsymbol{F}_{t}, \boldsymbol{F}_{s}\right)\right)\right)
\end{equation}
where~${ Concat}({\cdot})$~is channel-wise concatenation, $g(\cdot)$~is the convolution layer of size~$3\times 3$, and $\sigma(\cdot)$~is the~Sigmoid~activation function. The~$\boldsymbol{F}_t$ is adaptively superimposed into the $\boldsymbol{F}_s$ by soft gating~$\boldsymbol{G}_t$:
\begin{equation}
\boldsymbol{F}_{s}^{\prime}=\alpha\left(\boldsymbol{G}_{t} \odot \boldsymbol{F}_{t}\right) \oplus \boldsymbol{F}_{s}
\end{equation}
where $\alpha $ is a training parameter initialized to zero, $ \odot$ and $\oplus$ denote element-wise multiplication and element-wise addition, respectively. Symmetrically, we calculate the structure-aware texture
feature $\boldsymbol{F}_{t}^{\prime} $ as follows:
\begin{equation}
\boldsymbol{G}_{s}  =\sigma\left(h\left( { Concat }\left(\boldsymbol{F}_{t}, \boldsymbol{F}_{s}\right)\right)\right)
\end{equation}
\begin{equation}
\boldsymbol{F}_{t}^{\prime} =\beta\left(\boldsymbol{G}_{s} \odot \boldsymbol{F}_{s}\right) \oplus \boldsymbol{F}_{t}
\end{equation}
where $h$ follows the equivalent as $g$ and $\beta$ is a training parameter the same as $\alpha$.

\begin{figure*}[htbp]%
	\centering
	\includegraphics[width=0.95\textwidth]{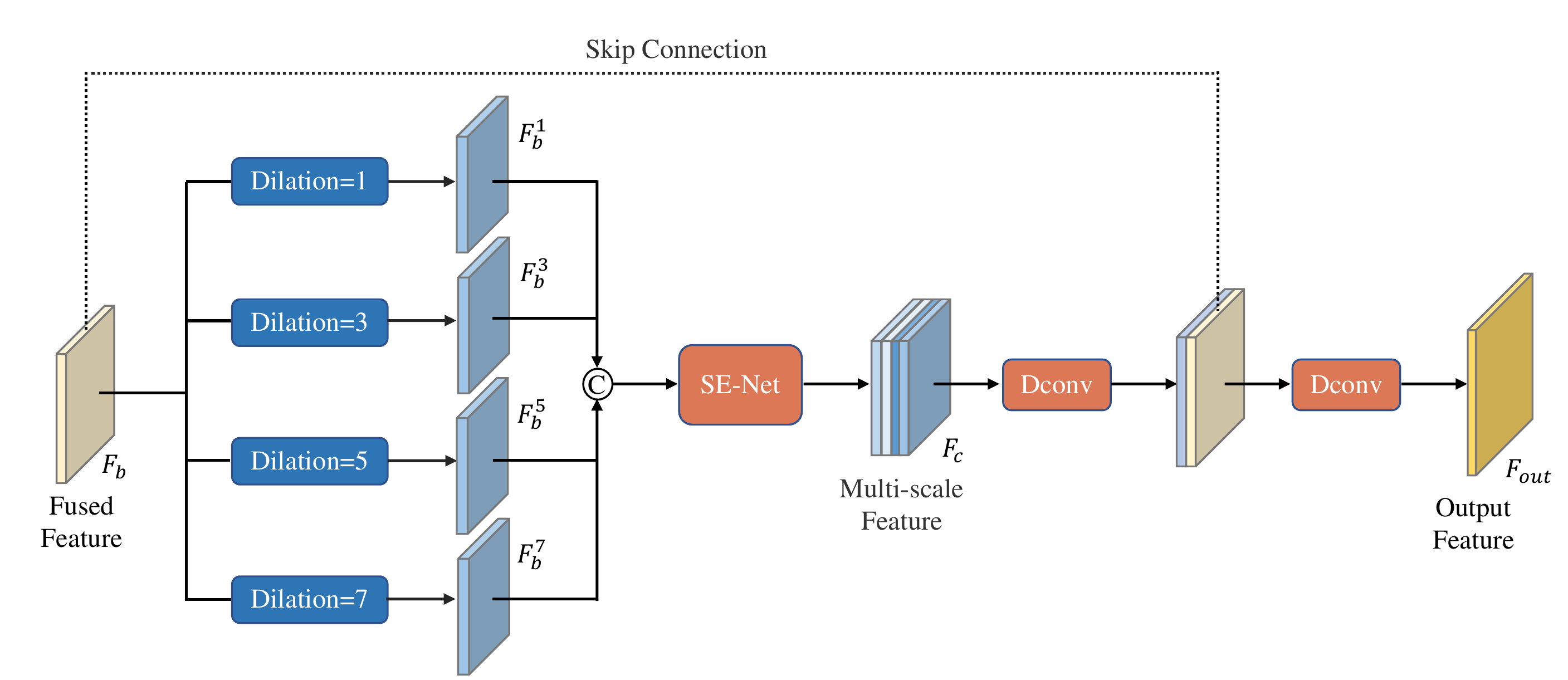}   %./figure/framework.eps
	\caption{The illustration of the multi-scale feature aggregation
(MFA) module, which owns the ability of dynamic selection from diverse semantic features of the murals, leading to semantic consistency results with adaptive features.}\label{model3}
\end{figure*}

Finally, we concatenate~$\boldsymbol{F}_s^{\prime}$~and~$\boldsymbol{F}_t^{\prime}$~in the channel dimension to obtain the fused feature~$\boldsymbol{F}_b$ after gated feature fusion:
\begin{equation}
\boldsymbol{F}_{b}={Concat}\left(\boldsymbol{F}_{s}^{\prime}, \boldsymbol{F}_{t}^{\prime}\right).
\end{equation}
\textbf{\subsection{Multi-scale feature aggregation}}
For better extracting beneficial features at various scales in the murals inpainting task, a multi-scale feature aggregation module is designed with multiple dilated convolutions for multi-scale semantics collection, as shown in Fig.~\ref{model3}.
\begin{equation}
\boldsymbol{F}_{b}^{k}=Conv_{k}\left(\boldsymbol{F}_{b}\right)
\end{equation}
where $\boldsymbol{F}_{b}$ is the fused feature from GFF module, and $Conv_{k} $~denotes dilated convolutional layers with expansion rate of~$k$, $k\in\{1,3,5,7\}$. We concatenate multi-scale semantic information on the channel dimension. Then, through a squeeze and excitation operations of SENet \cite{hu2018squeeze} to capture important channel information.
\begin{equation}
\boldsymbol{F}_{c}=SE({Concat}(\boldsymbol{F}_{b}^{1},\boldsymbol{F}_{b}^{3},\boldsymbol{F}_{b}^{5},\boldsymbol{F}_{b}^{7}))
\end{equation}

The skip connection is used to prevent semantic damage caused by patch-shift operations. The deconvolution layers are seamlessly embedded into our architecture to improve computational efficiency. The export $\boldsymbol{F}_{out}$ of the MFA module is represented as:
\begin{equation}
\boldsymbol{F}_{out}=Dconv({Concat}(Dconv(\boldsymbol{F}_{c}),\boldsymbol{F}_{b})).
\end{equation}
\textbf{\subsection{Loss functions}}
The model is trained with a joint loss, including reconstruction loss, perceptual loss, style loss, and adversarial loss, to render semantically reasonable and visually realistic results. In particular, for reconstruction loss, denote by $\boldsymbol{I}_{gt}$ the ground-truth image, initial binary mask $\boldsymbol{M}$ (with value 1 for the existing region, 0 otherwise),~$\boldsymbol{I}_{in}=\boldsymbol{I}_{gt} \odot \boldsymbol{M}$ the damaged image, the output of our generator is defined as $\boldsymbol{I}_{out}$, the reconstruction loss is defined as:
\begin{equation}
\mathcal{L}_{hole}=\frac{1}{N_{\boldsymbol{I}_{gt}}}\left\|(1-\boldsymbol{M}) \odot\left(\boldsymbol{I}_{out}-\boldsymbol{I}_{gt}\right)\right\|_{1}
\end{equation}
\begin{equation}
\mathcal{L}_{v a l i d}=\frac{1}{N_{\boldsymbol{I}_{gt}}}\left\|\boldsymbol{M} \odot\left(\boldsymbol{I}_{out}-\boldsymbol{I}_{gt}\right)\right\|_{1}
\end{equation}
where $N_{\boldsymbol{I}_{gt}} $ denotes the number of elements in $\boldsymbol{I}_{gt} $, the $\mathcal{L}_{hole}$ and $\mathcal{L}_{valid}$ are the reconstruction losses on the network output for the hole and the non-hole pixels respectively.

For capturing structure and texture information during generation learning, we introduce the perceptual loss, which measures the $l_1$ distance between $\boldsymbol{I}_{gt}$ and $ \boldsymbol{I}_{out}$ defined on the pre-trained VGG-16. Formally, let ${pool}_i$~is the feature map of the~$i^{th}$~th pooling layer of VGG-16. $H_i$, $W_i$, and $C_i$~denote the height, weight, and channel size of the~$i^{th}$~th feature map, respectively. The perceptual loss can be defined as:
\begin{equation}
\mathcal{L}_{ {perceptual }}=\sum_{i=1}^{N} \frac{1}{H_{i} W_{i} C_{i}}\left|\phi_{ {pool }_{i}}^{g t}-\phi_{ {pool }_{i}}^{ {out }}\right|_{1}
\end{equation}

We further introduce the style loss to preserve the style coherency. The style loss is similar to perceptual loss, which calculates as follows:
\begin{equation}
\psi_{ {pool }_{i}}=\phi_{ {pool }_{i}} \phi_{ {pool }_{i}}^{\mathsf{T}}
\end{equation}
\begin{equation}
\mathcal{L}_{{style }}=\sum_{i=1}^{N} \frac{1}{C_{i} \times C_{i}}\left|\frac{1}{H_{i} W_{i} C_{i}}\left(\psi_{\text {pool}_{i}}^{{{gt }}}-\psi_{\text {pool}_{i}}^{{ {out }}}\right)\right|_{1}
\end{equation}
where $\psi_{ {pool }_{i}}=\phi_{ {pool }_{i}} \phi_{ {pool }_{i}}^{\mathsf{T}}$ denotes the gram matrix constructed from the given feature maps. In summary, our joint loss function is:
\begin{equation}
\begin{aligned}
\mathcal{L}_{ {joint }}  =\lambda_{ {hole }} \mathcal{L}_{ {hole }}+\lambda_{ {valid }} \mathcal{L}_{ {valid }}+\lambda_{ {perceptual }} \mathcal{L}_{ {perceptual }}+\lambda_{ {style }} \mathcal{L}_{ {style }}
\end{aligned}
\end{equation}
for the tradeoff parameters, we empirically set 6 for $\lambda_{ {hole }}$, 1 for $\lambda_{ {valid }}$, 0.1 for $\lambda_{ {perceptual }}$, 180 for $\lambda_{ {style }}$.

\section{Experiments}
In this section, we present the experimental details, datasets, and comparison methods. Then, extensive experiments are conducted on our Dunhuang murals and two benchmark datasets for quantitative and qualitative evaluation. Finally, we perform ablation studies to validate the designed modules and innovative architecture of our model.

\textbf{\subsection{Training setting}}
The~model~is implemented using~pytorc~(1.12.0), with an i9-9700k CPU and~NVIDIA~RTX2080ti GPU. Train the model for 400,000 iterations with a learning rate of~1e-4, and finetuning for 200,000 iterations with a learning rate of~5e-5, using the Adam optimizer for network convergence, all masks and images are the size of~$256\times 256$.

\begin{figure*}[h]
	\centering
	\subfigcapskip=4pt % ?????????????
	\subfigure[Statue painting]{
		\begin{minipage}{0.23\linewidth}	
			\centering
			\includegraphics[width=1.05\linewidth]{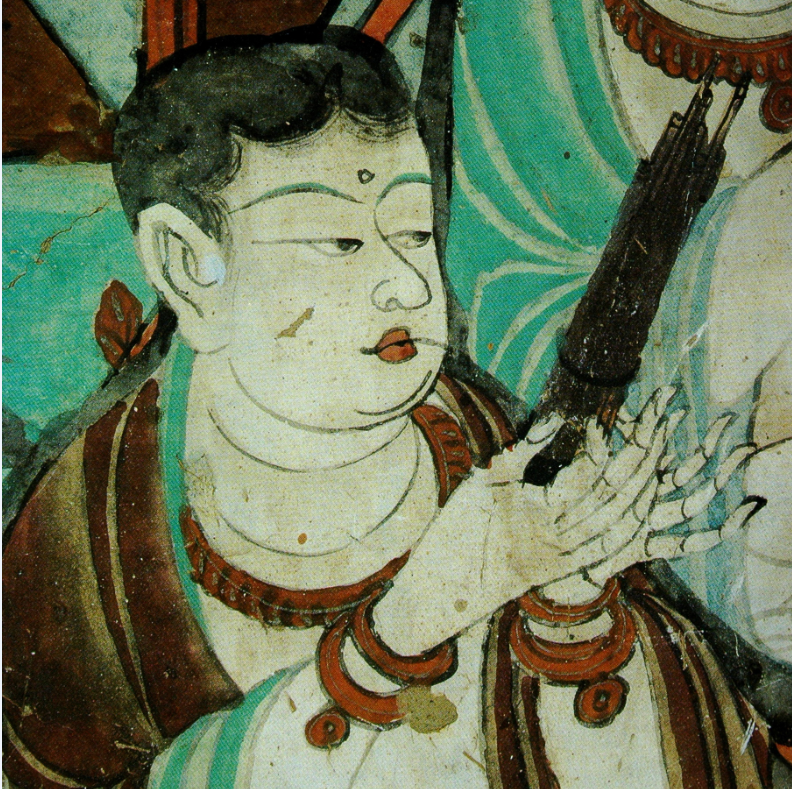}
		\end{minipage}
	}
	\subfigure[
Supernatural painting]{
		\begin{minipage}{0.23\linewidth}	
			\centering
			\includegraphics[width=1.05\linewidth]{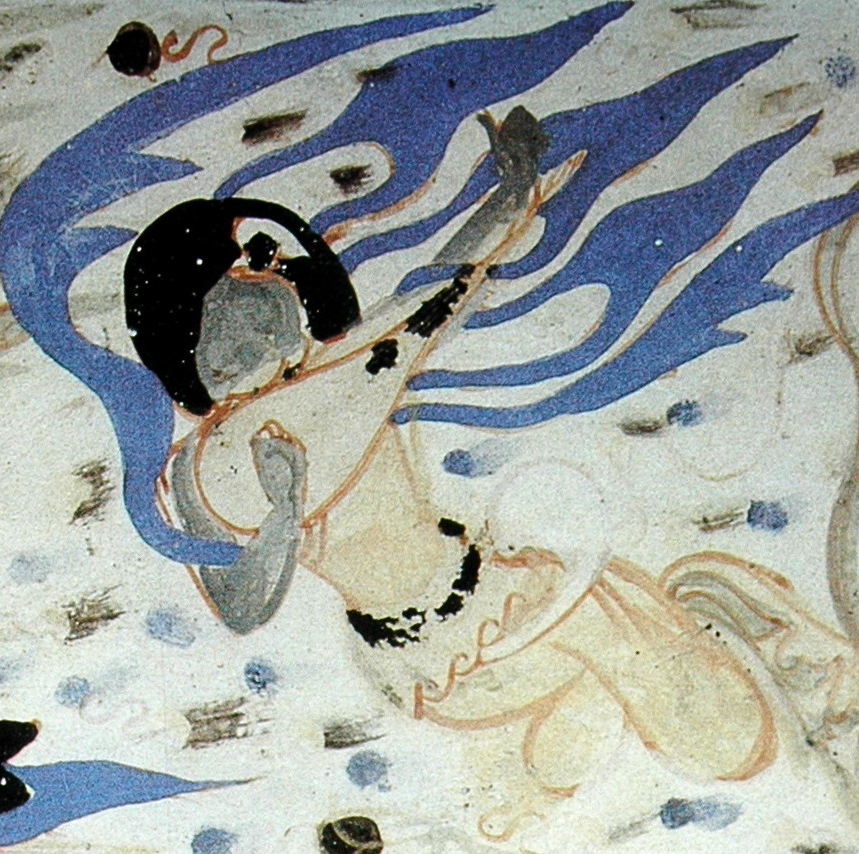}
		\end{minipage}
	}
	\subfigure[Sutra painting]{
		\begin{minipage}{0.23\linewidth}	
			\centering
			\includegraphics[width=1.05\linewidth]{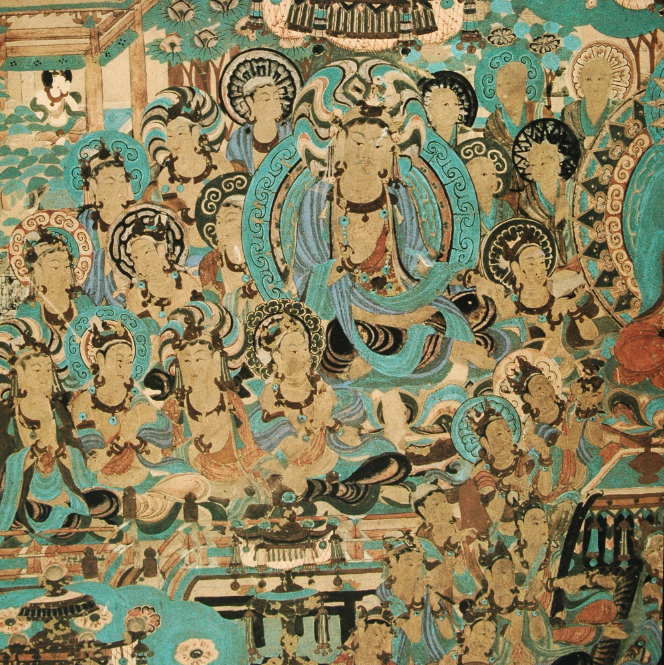}
		\end{minipage}
	}
	\subfigure[Support painting]{
		\begin{minipage}{0.23\linewidth}	
			\centering
			\includegraphics[width=1.05\linewidth]{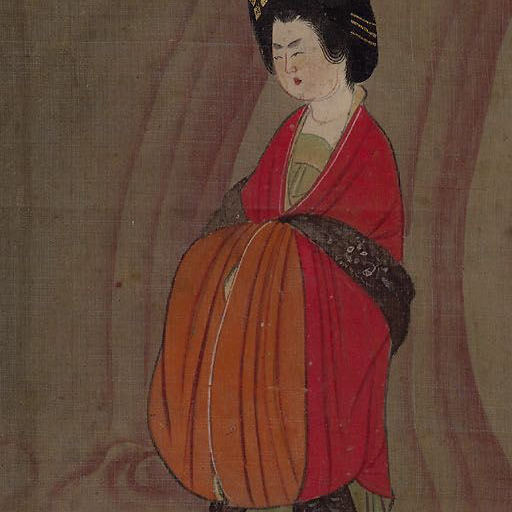}
		\end{minipage}
	}
	\hspace{0.1 cm}
	\caption{Dunhuang murals dataset.}
	\label{dunhuang3}
\end{figure*}

\textbf{\subsection{Datasets}}
$\textbf{Dunhuang murals dataset:}$~We constructed the Dunhuang murals dataset, which contains a training set of~5000~images and a testing set of~100~images mainly from the Mus¨¦e Guimet in France, the British Museum in the UK, the Dunhuang Research Institute, and partly obtained from high-resolution scans of Dunhuang murals publications. Fig.~\ref{dunhuang3}~shows different style paintings in the Dunhuang murals dataset.

$\textbf{Real masks dataset:}$~We extracted real masks from the broken regions of the mutilated murals are divided into~4~categories according to the breakage rate, with~1000~images in each category. The real masks can reflect the practical breakage characteristics of the mutilated murals.

$\textbf{Paris StreetView dataset:}$~A dataset consisting of 15000 images of the main buildings on the streets of Paris, with 14,900 images in the training set and 100 images in the testing set.

$\textbf{CelebA-HQ dataset:}$~Contains 30,000 high-resolution face images selected from the~CelebA~dataset, we randomly selected~29000~images as the training set and the remaining~1000~as test set.

\textbf{\subsection{Comparison models}}
We compare our method with five representative image inpainting methods in qualitative and quantitative experiments. The compared advanced methods are:
\begin{itemize}
     \item CA~\cite{yu2018generative}: An inpainting method using the contextual attention mechanism established the long-range dependencies of the feature map.
     \item GC~\cite{yu2019free}: A coarse-to-fine generative network contains gated convolution, which is proposed to implement a dynamic feature selection mechanism.
     \item LBAM~\cite{xie2019image}: An end-to-end learnable attention module is designated to accommodate irregular holes in convolutional layers.
     \item RFR~\cite{li2020recurrent}: An image inpainting method depends on recurrent feature reasoning with consistent attention for large hole regions.
     \item JPG~\cite{guo2021jpgnet}: A joint predictive method contains a generative network and filtering with an intelligent combination for murals inpainting.
\end{itemize}

\begin{figure*}[htbp]%
	\centering
	\includegraphics[width=\textwidth]{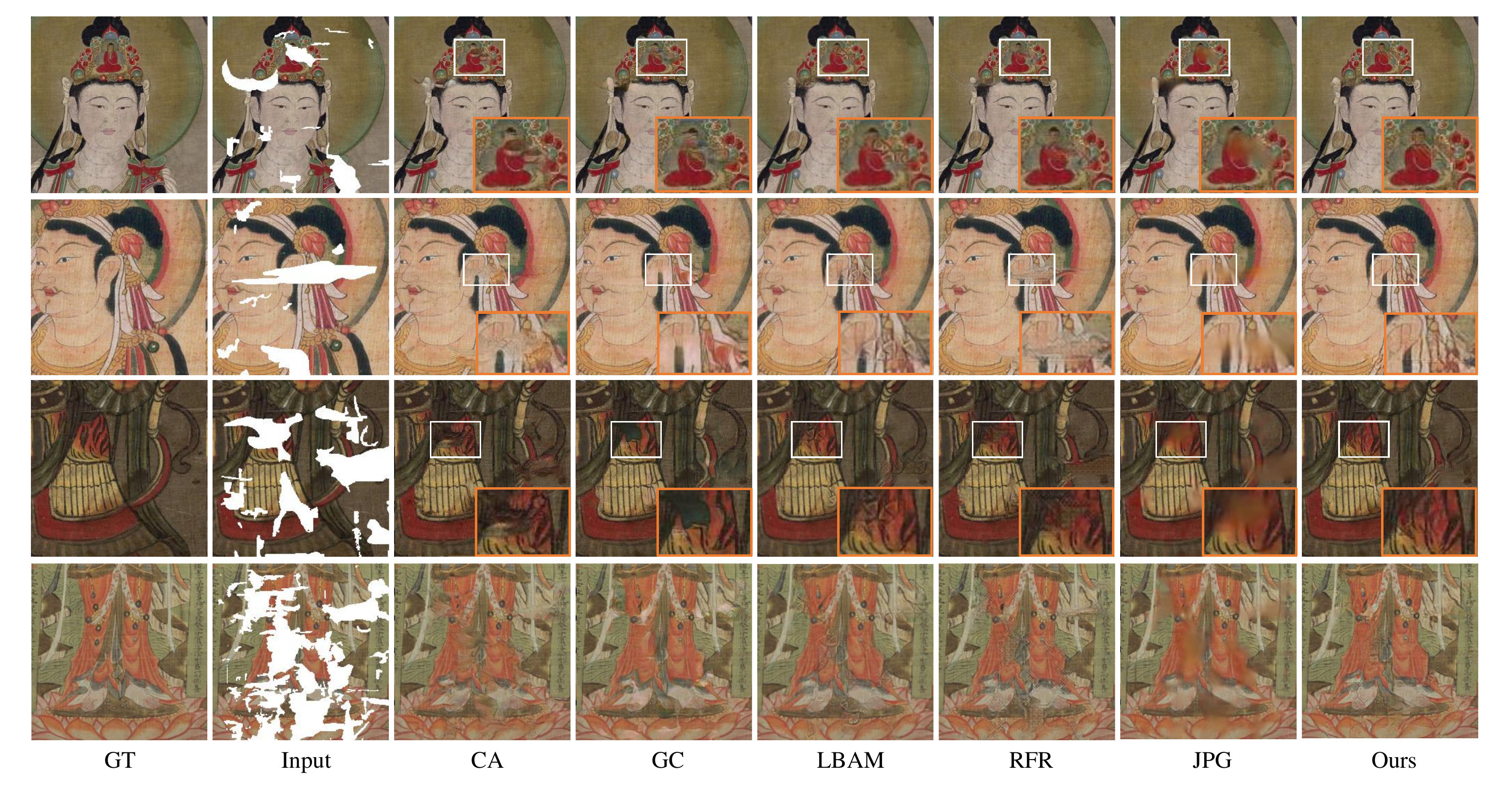}   %./figure/framework.eps
	\caption{Qualitative comparison results of our method with CA, GC, LBAM, RFR~and~JPG on the Dunhuang murals dataset using real masks.}\label{dunhuang}
\end{figure*}

\textbf{\subsection{Qualitative comparisons}}
Fig.~\ref{dunhuang}~demonstrates the inpainting results using real masks on the Dunhuang murals dataset. CA~and~LBAM~emerge different degrees of boundary blurring visually, while~GC and JPG~suffers from structural distortion and semantic discontinuities. The similar recurrent inpainting network~RFR~cannot obtains smooth inpainting results. In contrast, our inpainting results have a continuous global structure with precise texture details, which are visually more realistic. These results show that our MPR module fully uses the structure and texture information of the murals. In addition, the GFF and MFA modules effectively realize interaction and extraction of the structure and texture features, significantly enhancing the inference ability of the model.

\textbf{\subsection{Quantitative comparisons}}
For the evaluation metrics, we use several standard metrics in terms of picture signal-to-noise ratio peak (PSNR), structural similarity~(SSIM), fr¨¦chet inception distance (FID) and learned perceptual image patch similarity (LPIPS), where the first~2~metrics based on low-level pixel values and the latter~2~metrics associated with high levels of visual perception related. In the Dunhuang murals inpainting task, we divided the real mask into 4 groups by masking rate. For each category, we randomly use 100 masks to test the performance of our model. As shown in Table~\ref{dunhuangres}, our model achieves the best results for all evaluation metrics on~4~groups of masks.
\begin{table}[htbp]
\centering
\caption{Quantitative comparison results of different mask ratios on Dunhuang murals dataset, $\uparrow$ means higher is better, $\downarrow$ means lower is better, and bold indicates the best score.}
\begin{tabular}{c|l|c|c|c|c}
\toprule
\textbf{Metrics}~&~\textbf{Methods}~&~\textbf{(0.01--0.1]}~&~\textbf{(0.1--0.2]}~&~\textbf{(0.2--0.3]}~&~\textbf{(0.3--0.4]} \\
\midrule
  \multirow{6}{*}{PSNR$\uparrow$}          &CA~\cite{yu2018generative}   &   36.04 &  28.55 &24.67&22.39\\
            &GConv~\cite{yu2019free}  & 37.04 & 29.37& 25.32&22.71      \\
&{LBAM}~\cite{xie2019image} & 37.01 & 29.61& 25.83&23.76     \\
            &{RFR}~\cite{li2020recurrent} &36.46 & 28.83& 24.92&22.45     \\
            &{JPG}~\cite{guo2021jpgnet} &37.61 & 30.16& 26.33&23.97     \\
            &{OURS} & \textbf{37.86} & \textbf{30.21}& \textbf{26.41}&\textbf{24.23}    \\
\toprule
\multirow{6}{*}{SSIM$\uparrow$ } &CA~\cite{yu2018generative}   &   0.975 &  0.910 & 0.828&0.738\\
&GConv~\cite{yu2019free}  & 0.979 & 0.921& 0.843&0.749      \\
&{LBAM}~\cite{xie2019image} & 0.943 & 0.909& 0.792  &0.735  \\
&{RFR}~\cite{li2020recurrent} &0.966 & 0.912& 0.818    &0.725\\
&{JPG}~\cite{guo2021jpgnet} &0.985 & 0.938& 0.844    &0.759\\
&{OURS} & \textbf{0.986} & \textbf{0.942}& \textbf{0.853}  &\textbf{0.785}  \\
\toprule
 \multirow{6}{*}{FID$\downarrow$ }&CA~\cite{yu2018generative}   &   10.81 &  31.90 & 64.24&83.96\\
&GConv~\cite{yu2019free}  & 13.82 & 28.54& 58.48    &80.77  \\
 &{LBAM}~\cite{xie2019image} & 13.09 & 36.04& 63.72 &81.97   \\
&{RFR}~\cite{li2020recurrent} &13.69 & 41.44& 62.33   &81.02 \\
&{JPG}~\cite{guo2021jpgnet} &13.13 & 37.69& 78.48   &108.07 \\
&{OURS} & \textbf{10.45} & \textbf{27.84}& \textbf{55.59}  &\textbf{78.86}  \\
\toprule
 \multirow{6}{*}{LPIPS$\downarrow$ }   &CA~\cite{yu2018generative}   &   0.016 &  0.065 & 0.129&0.215\\
&GConv~\cite{yu2019free}  & 0.013 & 0.051& 0.109&0.189      \\
&{LBAM}~\cite{xie2019image} & 0.013 & 0.049& 0.112  &0.168  \\
&{RFR}~\cite{li2020recurrent} &0.016 & 0.062& 0.121 &0.199   \\
&{JPG}~\cite{guo2021jpgnet} &0.015 & 0.067& 0.139 &0.227   \\
&{OURS} & \textbf{0.011} & \textbf{0.044}& \textbf{0.104}   & \textbf{0.166}\\
\bottomrule
\end{tabular}
\label{dunhuangres}
\end{table}

\begin{figure*}[htbp]%
	\centering
	\includegraphics[width=\textwidth]{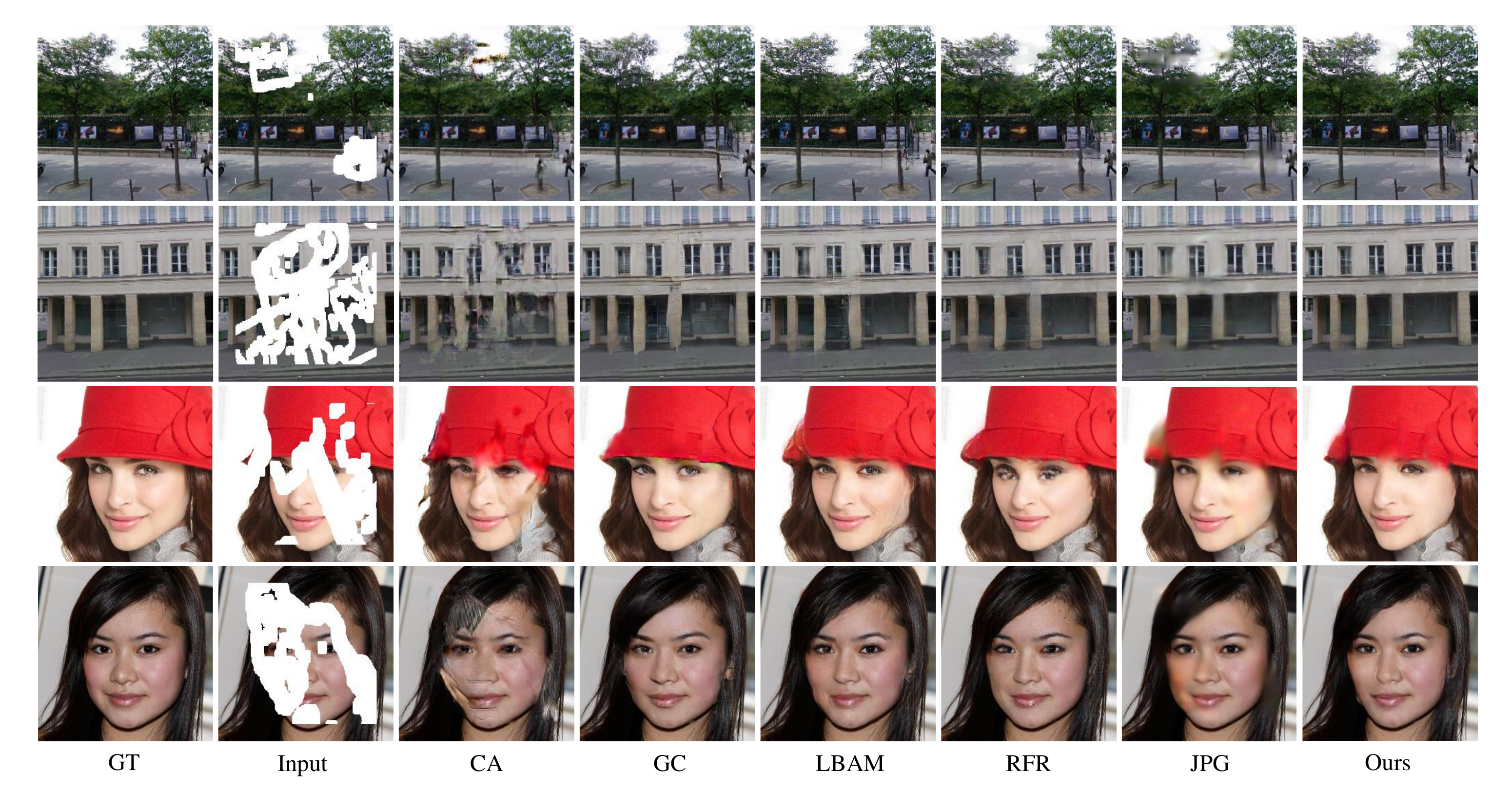}   %./figure/framework.eps
	\caption{Qualitative comparison results of our method with CA, GC, LBAM, RFR~and~JPG on the Paris StreetView and CelebA-HQ datasets using irregular masks.}\label{psvceleb}
\end{figure*}

\textbf{\subsection{Experiments on benchmark datasets}}
We conduct additional experiments to verify the generalization ability of the model on two benchmark datasets: Paris StreetView~and~CelebA-HQ, and the visualization results are shown in Fig.~\ref{psvceleb}. According to the observation, our proposed method can guarantee the semantic consistency of the restored content, with smooth and precise edges in the global structure and richer detail information in the local texture. Table~\ref{psvres}~shows the quantitative comparison results on the~Paris StreetView~dataset. Our model only slightly underperforms the~LBAM~method in~LPIPS~and~FID~metrics for small ratio masks and achieves the best inpainting results in all other comparison metrics. Overall, our method can maintain excellent inpainting performance in the quantitative and qualitative comparisons in benchmark datasets.
\begin{table}[htbp]
\centering
\caption{Quantitative comparison results of different mask ratios on Paris StreetView dataset, $\uparrow$ means higher is better, $\downarrow$ means lower is better, and bold indicates the best score.}
\begin{tabular}{c|l|c|c|c|c|c|c}
\toprule
\textbf{Metrics}&\textbf{Methods}& \textbf{(0.01--0.1]} & \textbf{(0.1--0.2]}&\textbf{(0.2--0.3]}&\textbf{(0.3--0.4]}& \textbf{(0.4--0.5]}&\textbf{(0.5--0.6]} \\
\midrule
\multirow{6}{*}{PSNR$\uparrow$} &CA~\cite{yu2018generative}   &   34.35 &  28.55 &25.67&23.39&21.22 &19.18\\
&GConv~\cite{yu2019free}  & 35.04 & 29.37& 26.32&24.51&22.45 &20.01      \\
&{LBAM}~\cite{xie2019image} &36.11	&30.57&	27.39&	25.23&	23.30&	20.91\\
&{RFR}~\cite{li2020recurrent} &36.36&	31.17&	28.08	&26.01	&24.15	&21.29\\
&{JPG}~\cite{guo2021jpgnet} &36.41&	31.34&	28.13	&25.94	&23.94	&21.26\\
&{OURS} & \textbf{37.04}	& \textbf{31.88}&	\textbf{28.69}&	\textbf{26.65}	&\textbf{24.76}	&\textbf{21.85} \\
\toprule
\multirow{6}{*}{SSIM$\uparrow$ }&CA~\cite{yu2018generative}  &   0.935 &  0.899 & 0.838&0.778&0.711 &0.627\\
&GConv~\cite{yu2019free}  & 0.959 & 0.921& 0.853&0.813  &0.741  &0.645    \\
 &{LBAM}~\cite{xie2019image} & 0.978	&0.936&	0.885	&0.827&	0.755	&0.657\\
&{RFR}~\cite{li2020recurrent} &0.977&	0.937	&0.889	&0.835&	0.770&	0.672\\
&{JPG}~\cite{guo2021jpgnet} &0.978&	0.939	&0.892	&0.836&	0.776&	0.692\\
&{OURS} & \textbf{0.986}&	\textbf{0.953}&	\textbf{0.894}	&\textbf{0.853}&	\textbf{0.791}&	\textbf{0.699}\\
\toprule
\multirow{6}{*}{FID$\downarrow$ }&CA~\cite{yu2018generative}  &   12.84 &  27.71 & 52.36&64.77&80.03 &121.79\\
&GConv~\cite{yu2019free}  & 10.82 & 25.54& 48.98    &58.77  &77.79 &99.91\\
 &{LBAM}~\cite{xie2019image} &\textbf{7.22}&	20.35&	35.67&	51.03&	69.72&	96.57  \\
&{RFR}~\cite{li2020recurrent} &9.93	&23.20&	42.27&	56.53	&82.07	&108.43\\
&{JPG}~\cite{guo2021jpgnet} &10.86	&26.52&	47.81&	70.72	&100.73	&149.32\\
&{OURS} & 8.93&	\textbf{19.73}&	\textbf{34.34}&	\textbf{47.18}&	\textbf{64.32}&	\textbf{83.25}\\
\toprule
 \multirow{6}{*}{LPIPS$\downarrow$ }&CA~\cite{yu2018generative}  &   0.021 &  0.067 & 0.099&0.143&0.197 &0.255\\
&GConv~\cite{yu2019free}  & 0.018 & 0.052& 0.089&0.129  & 0.175 &  0.268  \\
&{LBAM}~\cite{xie2019image} & \textbf{0.013}	&\textbf{0.041}&	\textbf{0.075}&	0.115&	0.168&	0.248 \\
&{RFR}~\cite{li2020recurrent} &0.016	&0.047&	0.084&	0.127	&0.181&	0.281\\
&{JPG}~\cite{guo2021jpgnet} &0.022	&0.065&	0.119&	0.179	&0.229&	0.304\\
&{OURS} & 0.015&	0.042&	\textbf{0.075}&	\textbf{0.112}	&\textbf{0.161}	&\textbf{0.242}\\
\bottomrule
\end{tabular}
\label{psvres}
\end{table}

\textbf{\subsection{Ablation studies}}
In this section, we would like to further validate our contributions. Here we conduct ablation experiments to research the influences of recurrences in TFR, and then implement several variants of our model to verify the effectiveness of the designed modules.

\begin{figure}[htbp]%
	\centering
	\includegraphics[width=0.93\textwidth]{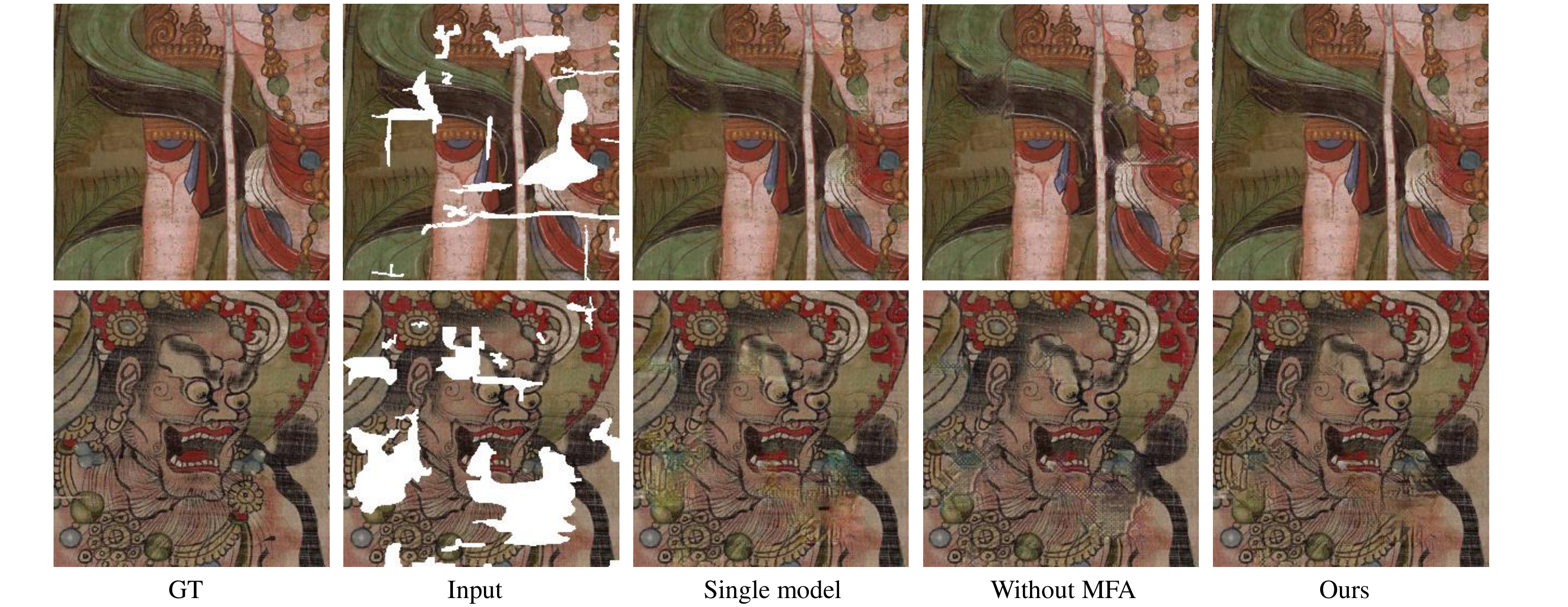}   %./figure/framework.eps
	\caption{Qualitative comparison results of controlled experiments.}\label{psvceleb2}
\end{figure}

\textbf{\subsubsection{The influences of iteration numbers}}
We fix 4 recurrences in SFR to ensure structural continuity and research the effect of iteration numbers in TFR on the Dunhuang murals dataset. The results corresponding to different iteration numbers are shown in Table \ref{iternum}. This ablation study reveals that our method owns robust, and the improved performance is from the more efficient architecture rather than the number of iterations.
\begin{table}[htbp]
\centering
\caption{The influences of different IterNums, the number indicates the iteration times in the second segment.}
\begin{tabular}{cccccc}
\toprule
\textbf{Metrics}&\textbf{IterNum}& \textbf{(0-0.1]} & \textbf{(0.1-0.2]}&\textbf{(0.2-0.3]}&\textbf{(0.3-0.4]} \\
\midrule
&3   &   37.56 &  30.02 &26.19&24.01\\
PSNR$\uparrow$ &{4} & \textbf{37.86} & \textbf{30.21}& \textbf{26.41}&\textbf{24.23}    \\
&{5} & {37.78} & {30.13}& {26.26} &{24.08}   \\
\toprule
&3   &   0.963 &  0.924 & 0.841&0.773\\
SSIM$\uparrow$ &{4} & \textbf{0.986} & \textbf{0.942}& \textbf{0.853}  &\textbf{0.785}   \\
&{5} & 0.981 &0.938& 0.848 &0.779 \\
\bottomrule
\end{tabular}
\label{iternum}
\end{table}
\textbf{\subsubsection{The effect of MPR and MFA modules}}
Previous recurrent inpainting methods tend to a single network architecture, which is challenging to satisfy the requirements of different receptive fields for the structure and texture in the mural inpainting tasks. To verify the superiority of the multi-stage architecture in our model, as the controlled experiment of a single model, we let the whole recurrent take place in the SFR under the same recurrences. In addition, we design a second set of control experiments, which lacks the MFA module compared to the original model. The qualitative and quantitative comparison results on the Dunhuang murals are shown in Fig.~\ref{psvceleb2}~and Table~\ref{xr2}.
\begin{table}[htbp]
\centering
\caption{Quantitative comparison results of the controlled experiments.}
\begin{tabular}{cccccc}
\toprule
\textbf{Metrics}&\textbf{Methods}& \textbf{(0-0.1]} & \textbf{(0.1-0.2]}&\textbf{(0.2-0.3]}&\textbf{(0.3-0.4]} \\
\midrule
&Single model   &   36.77 &  29.38 &25.89&23.73\\
PSNR$\uparrow$ &Without MFA &  37.62 &  29.97 &26.29&23.81  \\
&Ours & \textbf{37.86} & \textbf{30.21}& \textbf{26.41}&\textbf{24.23}    \\
\toprule
&Single model   &   0.934 &  0.911 & 0.839&0.767\\
SSIM$\uparrow$ &Without MFA & 0.976 &  0.939 & 0.850&0.779    \\
&Ours & \textbf{0.986} & \textbf{0.942}& \textbf{0.853}  &\textbf{0.785} \\
\bottomrule
\end{tabular}
\label{xr2}
\end{table}

\section{Conclusion}
\label{sec:conclusion}
In this paper, we propose a multi-stage progressive reasoning network for murals inpainting containing global to local receptive fields, enriching the information of missing regions and giving semantically explicit embedding results. In addition, a multi-scale feature aggregation module is designed to empower the capability of dynamic selection from significant features. The effectiveness of the proposed method is proved by conducting experiments on real damaged murals and comparison with five existing image inpainting methods. Additionally, we also studied the impact of mural outlines on mural inpainting tasks. In the future, we aim to further strengthen the constraints on the contour structure of the murals by introducing an edge extraction network. \\

\noindent \textbf{Conflicts of interest} The authors declare that they have no conflict of interest.\\

\noindent \textbf{Data availability statement} Data is openly available in a public repository.

\textbf{\section*{Acknowledgment}}
This work was supported in part by the Gansu Provincial Department of Education University Teachers Innovation Fund Project (No.2023B-056), the Introduction of Talent Research Project of Northwest Minzu University (No. xbmuyjrc201904), and the Fundamental Research Funds for the Central Universities of Northwest Minzu University (No.31920220019, 31920220037, 31920220130), the Leading Talent of National Ethnic Affairs Commission (NEAC), the Young Talent of NEAC, and the Innovative Research Team of NEAC (2018) 98.
\bibliographystyle{spbasic_unsort}%??????? ?????? ????????? ??????????.bib?

\bibliography{XJ_ref}

\end{document}